\documentclass[11pt]{article}
\usepackage{coling2016}
\usepackage{times}
\usepackage{url}
\usepackage{latexsym}

\usepackage{amsmath}
\usepackage{multirow}

\usepackage{graphicx}
\usepackage{subfigure}

\usepackage{amsfonts}

\title{Text Classification Improved by Integrating Bidirectional LSTM\\ with Two-dimensional Max Pooling}

\author{Peng Zhou$^{1}$, \ Zhenyu Qi$^{1}$\thanks{Correspondence author: zhenyu.qi@ia.ac.cn}, \ Suncong Zheng$^{1}$, \ Jiaming Xu$^{1}$, \ Hongyun Bao$^{1}$, \ Bo Xu$^{1,2}$ \\
  (1) Institute of Automation, Chinese Academy of Sciences, China \\(2) Center for Excellence in Brain Science and Intelligence Technology, China\\
  {\tt \{zhoupeng2013, zhenyu.qi, zhengsuncong,} \\
          {\tt jiaming.xu, hongyun.bao, xubo\}@ia.ac.cn}}

\date{}

\begin{document}
\maketitle
\begin{abstract}
Recurrent Neural Network (RNN) is one of the most popular architectures used in Natural Language Processsing (NLP) tasks because its recurrent structure is very suitable to process variable-length text. RNN can utilize distributed representations of words by first converting the tokens comprising each text into vectors, which form a matrix. And this matrix includes two dimensions: the time-step dimension and the feature vector dimension. Then most existing models usually utilize one-dimensional (1D) max pooling operation or attention-based operation only on the time-step dimension to obtain a fixed-length vector. 
However, the features on the feature vector dimension are not mutually independent, and simply applying 1D pooling operation over the time-step dimension independently may destroy the structure of the feature representation. On the other hand, applying two-dimensional (2D) pooling operation over the two dimensions may sample more meaningful features for sequence modeling tasks. To integrate the features on both dimensions of the matrix, this paper explores applying 2D max pooling operation to obtain a fixed-length representation of the text. This paper also utilizes 2D convolution to sample more meaningful information of the matrix. Experiments are conducted on six text classification tasks, including sentiment analysis, question classification, subjectivity classification and newsgroup classification. Compared with the state-of-the-art models, the proposed models achieve excellent performance on 4 out of 6 tasks. Specifically, one of the proposed models achieves highest accuracy on Stanford Sentiment Treebank binary classification and fine-grained classification tasks.



\end{abstract}

\section{Introduction}
\blfootnote{
    %
    %
    %
    %
    \hspace{-0.65cm}  
    This work is licenced under a Creative Commons 
    Attribution 4.0 International Licence.
    Licence details:
    \url{http://creativecommons.org/licenses/by/4.0/}
    %
    %
}

Text classification is an essential component in many NLP applications, such as sentiment analysis \cite{socher2013recursive}, relation extraction \cite{zeng2014relation} and spam detection \cite{wang2010don}. Therefore, it has attracted considerable attention from many researchers, and various types of models have been proposed. As a traditional method, the bag-of-words (BoW) model treats texts as unordered sets of words \cite{wang2012baselines}. In this way, however, it fails to encode word order and syntactic feature.

Recently, order-sensitive models based on neural networks have achieved tremendous success for text classification, and shown more significant progress compared with BoW models. The challenge for textual modeling is how to capture features for different text units, such as phrases, sentences and documents. Benefiting from its recurrent structure, RNN, as an alternative type of neural networks, is very suitable to process the variable-length text.

RNN can capitalize on distributed representations of words by first converting the tokens comprising each text into vectors, which form a matrix. This matrix includes two dimensions: the time-step dimension and the feature vector dimension, and it will be updated in the process of learning feature representation. 
Then RNN utilizes 1D max pooling operation \cite{lai2015recurrent} or attention-based operation \cite{zhou2016attention}, which extracts maximum values or generates a weighted representation over the time-step dimension of the matrix, to obtain a fixed-length vector. Both of the two operators ignore features on the feature vector dimension, which maybe important for sentence representation, therefore the use of 1D max pooling and attention-based operators may pose a serious limitation. 



Convolutional Neural Networks (CNN) \cite{kalchbrenner2014convolutional,kim2014convolutional} utilizes 1D convolution to perform the feature mapping, and then applies 1D max pooling operation over the time-step dimension to obtain a fixed-length output. However the elements in the matrix learned by RNN are not independent, as RNN reads a sentence word by word, one can effectively treat the matrix as an 'image'. Unlike in NLP, CNN in image processing tasks \cite{lecun1998gradient,krizhevsky2012imagenet} applies 2D convolution and 2D pooling operation to get a representation of the input. It is a good choice to utilize 2D convolution and 2D pooling to sample more meaningful features on both the time-step dimension and the feature vector dimension for text classification.

Above all, this paper proposes Bidirectional Long Short-Term Memory Networks with Two-Dimensional Max Pooling (BLSTM-2DPooling) to capture features on both the time-step dimension and the feature vector dimension. It first utilizes Bidirectional Long Short-Term Memory Networks (BLSTM) to transform the text into vectors. And then 2D max pooling operation is utilized to obtain a fixed-length vector. This paper also applies 2D convolution (BLSTM-2DCNN) to capture more meaningful features to represent the input text. 

The contributions of this paper can be summarized as follows:

\begin{itemize}
\item This paper proposes a combined framework, which utilizes BLSTM to capture long-term sentence dependencies, and extracts features by 2D convolution and 2D max pooling operation for sequence modeling tasks. To the best of our knowledge, this work is the first example of using 2D convolution and 2D max pooling operation in NLP tasks.

\item This work introduces two combined models BLSTM-2DPooling and BLSTM-2DCNN, and verifies them on six text classification tasks, including sentiment analysis, question classification, subjectivity classification, and newsgroups classification. Compared with the state-of-the-art models, BLSTM-2DCNN achieves excellent performance on $4$ out of $6$ tasks. Specifically, it achieves highest accuracy on Stanford Sentiment Treebank binary classification and fine-grained classification tasks.

\item To better understand the effect of 2D convolution and 2D max pooling operation, this paper conducts experiments on Stanford Sentiment Treebank fine-grained task. It first depicts the performance of the proposed models on different length of sentences, and then conducts a sensitivity analysis of 2D filter and max pooling size. 
\end{itemize}

The remainder of the paper is organized as follows. In Section 2, the related work about text classification is reviewed. Section 3 presents the BLSTM-2DCNN architectures for text classification in detail. Section 4 describes details about the setup of the experiments. Section 5 presents the experimental results. The conclusion is drawn in the section 6.

\section{Related Work}
Deep learning based neural network models have achieved great improvement on text classification tasks. These models generally consist of a projection layer that maps words of text to vectors. And then combine the vectors with different neural networks to make a fixed-length representation. According to the structure, they may divide into four categories: Recursive Neural Networks (RecNN\footnote{To avoid confusion with RNN, we named Recursive Neural Networks as RecNN.}), RNN, CNN and other neural networks.

\textbf{Recursive Neural Networks}: RecNN is defined over recursive tree structures. In the type of recursive models, information from the leaf nodes of a tree and its internal nodes are combined in a bottom-up manner. \newcite{socher2013recursive} introduced recursive neural tensor network to build representations of phrases and sentences by combining neighbour constituents based on the parsing tree. \newcite{irsoy2014deep} proposed deep recursive neural network, which is constructed by stacking multiple recursive layers on top of each other, to modeling sentence.

\textbf{Recurrent Neural Networks}: RNN has obtained much attention because of their superior ability to preserve sequence information over time. \newcite{tang2015target} developed target dependent Long Short-Term Memory Networks (LSTM \cite{hochreiter1997long}), where target information is automatically taken into account. \newcite{tai2015improved} generalized  LSTM to Tree-LSTM where each LSTM unit gains information from its children units. \newcite{zhou2016attention} introduced BLSTM with attention mechanism to automatically select features that have a decisive effect on classification. \newcite{yang2016hierarchical} introduced a hierarchical network with two levels of attention mechanisms, which are word attention and sentence attention, for document classification. This paper also implements an attention-based model BLSTM-Att like the model in \newcite{zhou2016attention}.



\textbf{Convolution Neural Networks}: CNN \cite{lecun1998gradient} is a feedforward neural network with 2D convolution layers and 2D pooling layers, originally developed for image processing. Then CNN is applied to NLP tasks, such as sentence classification \cite{kalchbrenner2014convolutional,kim2014convolutional}, and relation classification \cite{zeng2014relation}. The difference is that the common CNN in NLP tasks is made up of 1D convolution layers and 1D pooling layers. \newcite{kim2014convolutional} defined a CNN architecture with two channels. \newcite{kalchbrenner2014convolutional} proposed a dynamic $k$-max pooling mechanism for sentence modeling. \cite{zhang2015sensitivity} conducted a sensitivity analysis of one-layer CNN to explore the effect of architecture components on model performance. \newcite{yin2016multichannel} introduced multichannel embeddings and unsupervised pretraining to improve classification accuracy. \cite{zhang2015sensitivity} conducted a sensitivity analysis of one-layer CNN to explore the effect of architecture components on model performance.

Usually there is a misunderstanding that 1D convolutional filter in NLP tasks has one dimension. Actually it has two dimensions $(k, d)$, where $k$, $d \in \mathbb{R}$. As $d$ is equal to the word embeddings size $d^w$, the window slides only on the time-step dimension, so the convolution is usually called 1D convolution. While $d$ in this paper varies from 2 to $d^w$, to avoid confusion with common CNN, the convolution in this work is named as 2D convolution. The details will be described in Section~\ref{cnn_detail}.




\textbf{Other Neural Networks}: In addition to the models described above, lots of other neural networks have been proposed for text classification. \newcite{iyyer2015deep} introduced a deep averaging network, which fed an unweighted average of word embeddings through multiple hidden layers before classification. \newcite{zhou2015c} used CNN to extract a sequence of higher-level phrase representations, then the representations were fed into a LSTM to obtain the sentence representation.


The proposed model BLSTM-2DCNN is most relevant to DSCNN \cite{zhang2016dependency} and RCNN \cite{wen2016learning}. The difference is that the former two utilize LSTM, bidirectional RNN respectively, while this work applies BLSTM, to capture long-term sentence dependencies. After that the former two both apply 1D convolution and 1D max pooling operation, while this paper uses 2D convolution and 2D max pooling operation, to obtain the whole sentence representation.


\section{Model}

\begin{figure*}[t]
    \centering
    \includegraphics[width=1.\linewidth]{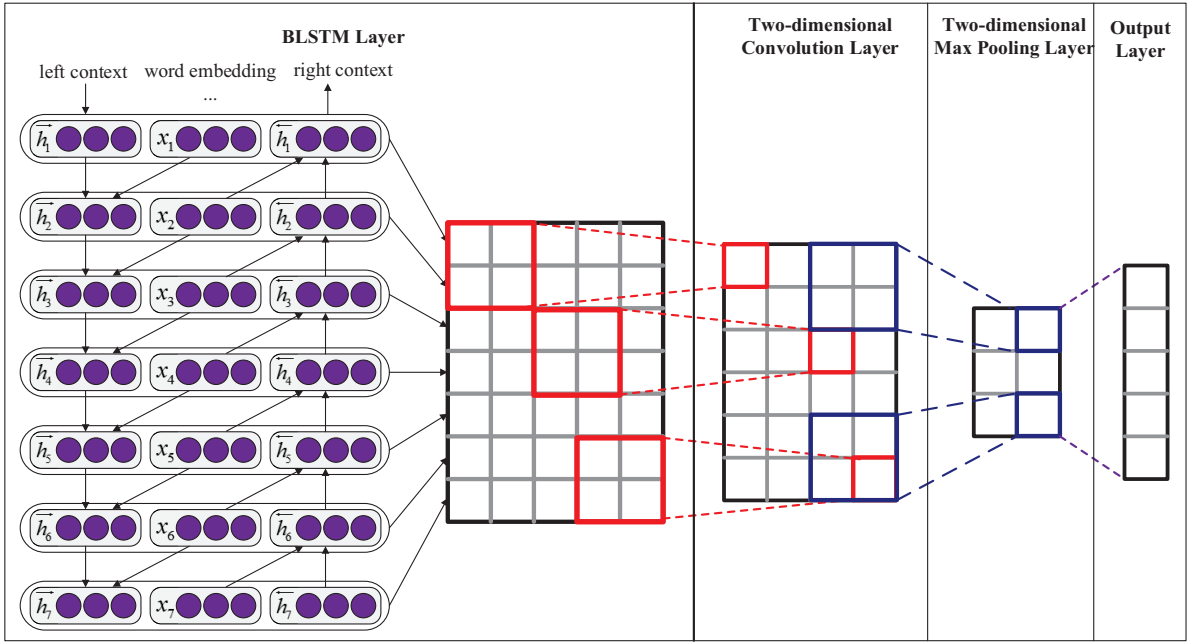}
    \caption{A BLSTM-2DCNN for the seven word input sentence. Word embeddings have size 3, and BLSTM has 5 hidden units. The height and width of convolution filters and max pooling operations are 2, 2 respectively.}
\end{figure*}

As shown in Figure 1, the overall model consists of four parts: BLSTM Layer, Two-dimensional Convolution Layer, Two dimensional max pooling Layer, and Output Layer. The details of different components are described in the following sections.
\subsection{BLSTM Layer}

LSTM was firstly proposed by \newcite{hochreiter1997long} to overcome the gradient vanishing problem of RNN. The main idea is to introduce an adaptive gating mechanism, which decides the degree to keep the previous state and memorize the extracted features of the current data input. 
Given a sequence $S =\{x_1, x_2, \dots, x_l\}$, where $l$ is the length of input text, LSTM processes it word by word. At time-step $t$, the memory $c_t$ and the hidden state $h_t$ are updated with the following equations:

\begin{equation}
\begin{bmatrix}
i_t\\f_t\\o_t\\\hat{c}_t
\end{bmatrix}
 = 
\begin{bmatrix}
\sigma\\ \sigma\\ \sigma\\ \tanh
\end{bmatrix}
W\cdot \lbrack h_{t-1}, x_t \rbrack\\
\end{equation}

\begin{eqnarray}
c_t & = & f_t \odot c_{t-1} + i_t \odot \hat{c}_t\\
h_t & = & o_t \odot \tanh(c_t)
\end{eqnarray}
where $x_t$ is the input at the current time-step, $i$, $f$ and $o$ is the input gate activation, forget gate activation and output gate activation respectively, $\hat{c}$ is the current cell state, $\sigma$ denotes the logistic sigmoid function and $\odot$ denotes element-wise multiplication.

For the sequence modeling tasks, it is beneficial to have access to the past context as well as the future context. \newcite{schuster1997bidirectional} proposed BLSTM to extend the unidirectional LSTM by introducing a second hidden layer, where the hidden to hidden connections flow in opposite temporal order. Therefore, the model is able to exploit information from both the past and the future.

In this paper, BLSTM is utilized to capture the past and the future information. As shown in Figure 1, the network contains two sub-networks for the forward and backward sequence context respectively. 
The output of the $i^{th}$ word is shown in the following equation:
\begin{equation}
h_i=[\overrightarrow{h_i} \oplus \overleftarrow{h_i}]
\end{equation}

Here, element-wise sum is used to combine the forward and backward pass outputs.

\subsection{Convolutional Neural Networks}
\label{cnn_detail}
Since BLSTM has access to the future context as well as the past context, $h_i$ is related to all the other words in the text. One can effectively treat the matrix, which consists of feature vectors, as an 'image', so 2D convolution and 2D max pooling operation can be utilized to capture more meaningful information. 
\subsubsection{Two-dimensional Convolution Layer}
A matrix $H = \{h_1, h_2, \dots, h_l\}$, $H \in \mathbb{R}^{l \times d^w}$, is obtained from BLSTM Layer, where $d^w$ is the size of word embeddings. 
Then narrow convolution is utilized \cite{kalchbrenner2014convolutional} to extract local features over $H$. A convolution operation involves a 2D filter $\mathbf{m} \in \mathbb{R}^{k \times d}$, which is applied to a window of k words and d feature vectors. For example, a feature $o_{i, j}$ is generated from a window of vectors $H_{i:i+k-1, \; j:j+d-1}$ by
\begin{equation}
o_{i, j} = f(\mathbf{m}  \cdot H_{i:i+k-1, \; j:j+d-1} + b)
\end{equation}
where $i$ ranges from 1 to $(l-k+1)$, $j$ ranges from 1 to $(d^w-d+1)$, $\cdot$ represents dot product, $b \in \mathbb{R}$ is a bias and an $f$ is a non-linear function such as the hyperbolic tangent. This filter is applied to each possible window of the matrix $H$ to produce a feature map $O$:
\begin{equation}
O = [o_{1,1}, o_{1,2}, \cdots, o_{l-k+1, d^w-d+1}] 
\end{equation}
with $O \in \mathbb{R}^{(l-k+1) \times (d^w-d+1)}$. It has described the process of one convolution filter. The convolution layer may have multiple filters for the same size filter to learn complementary features, or multiple kinds of filter with different size.

\subsubsection{Two-dimensional Max Pooling Layer}

Then 2D max pooling operation is utilized to obtain a fixed length vector. For a 2D max pooling $p \in \mathbb{R}^{p_1 \times p_2}$, it is applied to each possible window of matrix O to extract the maximum value:
\begin{equation}
p_{i,j} = down{\left(O_{i:i+p_1, \; j:j+p_2}\right)}
\end{equation}
where $down(\cdot)$ represents the 2D max pooling function, $i= (1, 1+p_1,  \cdots,  1+(l-k+1/p_1-1) \cdot p_1)$,  and $j = (1, \; 1+p_2, \cdots, 1+(d^w-d+1/p_2-1) \cdot p_2)$.
Then the pooling results are combined as follows:
\begin{equation}
h^* = [p_{1, 1},  p_{1,  1+p_2}, \cdots, p_{1+(l-k+1/p_1-1) \cdot p_1, 1+(d^w-d+1/p_2-1) \cdot p_2}]
\end{equation}
where $h^* \in \mathbb{R}$, and the length of $h^*$ is $\lfloor l-k+1/p_1 \rfloor \times \lfloor d^w-d+1/p_2 \rfloor$.


\subsection{Output Layer}
For text classification, the output $h^*$ of 2D Max Pooling Layer is the whole representation of the input text $S$. And then it is passed to a softmax classifier layer to predict the semantic relation label $\hat{y}$ from a discrete set of classes $\mathit{Y}$. The classifier takes the hidden state $h^*$ as input:
\begin{eqnarray}
\hat{p}\left(y | s\right) & = & softmax \left(W^{\left(s\right)} h^* + b^{\left(s\right)}\right)\\
\hat{y} &  = & \arg \max_y \hat{p}\left(y | s\right)
\end{eqnarray}

A reasonable training objective to be minimized is the categorical cross-entropy loss.
The loss is calculated as a regularized sum:

\begin{equation}
J\left(\theta\right) = -\frac{1}{m}\sum_{i=1}^{m}t_i\log(y_i)  + \lambda{\Vert\theta\Vert}_F^2
\end{equation}
where $\boldsymbol{t} \in \mathbb{R}^m$ is the one-hot represented ground truth, $\boldsymbol{y} \in \mathbb{R}^m$ is the estimated probability for each class by softmax, $m$ is the number of target classes, and $\lambda$ is an L2 regularization hyper-parameter. Training is done through stochastic gradient descent over shuffled mini-batches with the AdaDelta \cite{zeiler2012adadelta} update rule. Training details are further introduced in Section~\ref{hyper-parameter}.



\begin{table*}[!t]
\centering
\begin{tabular}{c||c|c|c|c|c|c|c|c}
\hline
{\bf{Data}} & {c} & {l} & {m} & {train} & {dev} & {test} & {$|V|$} & {$|V_{pre}|$} \\

\hline
SST-1    &5    &18    &51    &8544    &1101    &2210    &17836    &12745\\
SST-2    &2    &19    &51    &6920    &872    &1821    &16185    &11490\\
Subj    &2    &23    &65    &10000    &-    &\textbf{CV}    &21057    &17671\\
TREC    &6    &10    &33    &5452    &-    &500    &9137    &5990\\
MR    &2    &21    &59    &10662    &-    &\textbf{CV}    &20191    &16746\\
20Ng    &4    &276    &11468    &7520    &836    &5563    &51379    &30575\\
\hline
\end{tabular}
\caption{Summary statistics for the datasets. c: number of target classes, l: average sentence length, m: maximum sentence length, train/dev/test: train/development/test set size, $|V|$: vocabulary size, $|V_{pre}|$: number of words present in the set of pre-trained word embeddings, \textbf{CV}: 10-fold cross validation.}\label{tab:result}
\end{table*}

\section{Experimental Setup}
\subsection{Datasets}
The proposed models are tested on six datasets. Summary statistics of the datasets are in Table 1.
\begin{itemize}
\item \textbf{MR}\footnote{https://www.cs.cornell.edu/people/pabo/movie-review-data/}:    Sentence polarity dataset from \newcite{pang2005seeing}. The task is to detect positive/negative reviews.

\item \textbf{SST-1}\footnote{http://nlp.stanford.edu/sentiment/}:   Stanford Sentiment Treebank is an extension of MR from \newcite{socher2013recursive}. The aim is to classify a review as fine-grained labels (very negative, negative, neutral, positive, very positive).

\item \textbf{SST-2}:    Same as SST-1 but with neutral reviews removed and binary labels (negative, positive). For both experiments, phrases and sentences are used to train the model, but only sentences are scored at test time \cite{socher2013recursive,le2014distributed}. Thus the training set is an order of magnitude larger than listed in table 1.

\item \textbf{Subj}\footnote{http://www.cs.cornell.edu/people/pabo/movie-review-data/}:    Subjectivity dataset \cite{pang2004sentimental}. The task is to classify a sentence as being subjective or objective.

\item \textbf{TREC}\footnote{http://cogcomp.cs.illinois.edu/Data/QA/QC/}:    Question classification dataset \cite{li2002learning}. The task involves classifying a question into 6 question types (abbreviation, description, entity, human, location, numeric value).

\item \textbf{20Newsgroups}\footnote{http://web.ist.utl.pt/acardoso/datasets/}:    The 20Ng dataset contains messages from twenty newsgroups. We use the bydate version preprocessed by \newcite{cachopo2007improving}. We select four major categories (comp, politics, rec and religion) followed by \newcite{hingmire2013document}. 
\end{itemize}


\subsection{Word Embeddings}
The word embeddings are pre-trained on much larger unannotated corpora to achieve better generalization given limited amount of training data \cite{turian2010word}. In particular, our experiments utilize the GloVe embeddings\footnote{http://nlp.stanford.edu/projects/glove/} trained by \newcite{pennington2014glove} on 6 billion tokens of Wikipedia 2014 and Gigaword 5. Words not present in the set of pre-trained words are initialized by randomly sampling from uniform distribution in $[-0.1, 0.1]$. The word embeddings are fine-tuned during training to improve the performance of classification.

\subsection{Hyper-parameter Settings}
\label{hyper-parameter}
For datasets without a standard development set we randomly select $10\%$ of the training data as the development set. The evaluation metric of the 20Ng is the Macro-F1 measure followed by the state-of-the-art work and the other five datasets use accuracy as the metric. 
The final hyper-parameters are as follows. 

The dimension of word embeddings is 300, the hidden units of LSTM is 300. We use 100 convolutional filters each for window sizes of (3,3), 2D pooling size of (2,2). We set the mini-batch size as 10 and the learning rate of AdaDelta as the default value 1.0.
For regularization, we employ Dropout operation \cite{hinton2012improving} with dropout rate of 0.5 for the word embeddings, 0.2 for the  BLSTM layer and 0.4 for the penultimate layer, we also use l2 penalty with coefficient $10^{-5}$ over the parameters.

These values are chosen via a grid search on the SST-1 development set. We only tune these hyper-parameters, and more finer tuning, such as using different numbers of hidden units of LSTM layer, or using wide convolution \cite{kalchbrenner2014convolutional}, may further improve the performance.

\begin{table*}[!t]
\centering
\begin{tabular}{c||l|c|c|c|c|c|c}
\hline
{\bf{NN}} & {\bf{Model}} & {\bf{SST-1}} & {\bf{SST-2}} & {\bf{Subj}} & {\bf{TREC}} & {\bf{MR}} & {\bf{20Ng}} \\
\hline
\multirow{2}{*}{ReNN}
&RNTN \cite{socher2013recursive}    &45.7    &85.4    &-    &-    &-    &-\\
&DRNN \cite{irsoy2014deep}    &49.8    &86.6    &-    &-    &-    &-\\
\hline
\multirow{8}{*}{CNN}
&DCNN \cite{kalchbrenner2014convolutional}    &48.5    &86.8    &-    &93.0    &-    &-\\
&CNN-non-static \cite{kim2014convolutional}    &48.0    &87.2    &93.4    &93.6    &-    &-\\
&CNN-MC \cite{kim2014convolutional}    &47.4    &88.1    &93.2    &92    &-    &-\\
&TBCNN\cite{mou2015discriminative}    &51.4    &87.9    &-    &96.0    &-    &-\\
&Molding-CNN \cite{lei2015molding} &51.2   &88.6    &-    &-    &-   &-\\
&CNN-Ana \cite{zhang2015sensitivity}    &45.98    &85.45    &93.66    &91.37    &81.02    &-\\
&MVCNN \cite{yin2016multichannel}    &49.6    &89.4    &93.9    &-    &-   &-\\  
\hline
\multirow{7}{*}{RNN}
&RCNN \cite{lai2015recurrent}     &47.21    &-    &-    &-    &-   &96.49\\
&S-LSTM \cite{zhu2015long}    &-    &81.9    &-    &-    &-   &-\\
&LSTM \cite{tai2015improved}    &46.4    &84.9    &-    &-    &-   &-\\
&BLSTM \cite{tai2015improved}     &49.1    &87.5     &-    &-    &-   &-\\
&Tree-LSTM \cite{tai2015improved}    &51.0    &88.0    &-    &-    &-   &-\\
&LSTMN \cite{cheng2016long}    &49.3    &87.3    &-    &-    &-   &-\\
&Multi-Task \cite{liu2016recurrent}    &49.6    &87.9    &94.1    &-    &-   &-\\
\hline
\multirow{7}{*}{Other}
&PV \cite{le2014distributed}    &48.7    &87.8    &-    &-    &-   &-\\
&DAN \cite{iyyer2015deep}    &48.2    &86.8    &-    &-    &-   &-\\
&combine-skip \cite{kiros2015skip}    &-    &-    &93.6    &92.2    &76.5    &-\\
&AdaSent \cite{zhao2015self}    &-    &-    &\textbf{95.5}    &92.4    &\textbf{83.1}    &-\\
&LSTM-RNN \cite{le2015compositional}    &49.9    &88.0    &-    &-    &-   &-\\
&C-LSTM \cite{zhou2015c}    &49.2    &87.8    &-    &94.6    &-   &-\\
&DSCNN \cite{zhang2016dependency}    &49.7    &89.1    &93.2    &95.4    &81.5    &-\\
\hline
\multirow{4}{*}{ours}
&BLSTM    &49.1    &87.6    &92.1    &93.0    &80.0    &94.0\\
&BLSTM-Att    &49.8    &88.2    &93.5    &93.8    &81.0    &94.6\\
&BLSTM-2DPooling    &50.5    &88.3    &93.7    &94.8    &81.5    &95.5\\
&BLSTM-2DCNN    &\textbf{52.4}    &\textbf{89.5}    &94.0    &\textbf{96.1}   &82.3     &\textbf{96.5}\\
\hline


\end{tabular}
\caption{Classification results on several standard benchmarks. \textbf{RNTN}: Recursive deep models for semantic compositionality over a sentiment treebank \protect\cite{socher2013recursive}. \textbf{DRNN}: Deep recursive neural networks for compositionality in language \protect\cite{irsoy2014deep}. \textbf{DCNN}: A convolutional neural network for modeling sentences \protect\cite{kalchbrenner2014convolutional}. \textbf{CNN-nonstatic/MC}: Convolutional neural networks for sentence classification \protect\cite{kim2014convolutional}. \textbf{TBCNN}: Discriminative neural sentence modeling by tree-based convolution \protect\cite{mou2015discriminative}. \textbf{Molding-CNN}: Molding CNNs for text: non-linear, non-consecutive convolutions \protect\cite{lei2015molding}. \textbf{CNN-Ana}: A Sensitivity Analysis of (and Practitioners' Guide to) Convolutional Neural Networks for Sentence Classification \protect\cite{zhang2015sensitivity}. \textbf{MVCNN}: Multichannel variable-size convolution for sentence classification \protect\cite{yin2016multichannel}. \textbf{RCNN}: Recurrent Convolutional Neural Networks for Text Classification \protect\cite{lai2015recurrent}. \textbf{S-LSTM}: Long short-term memory over recursive structures \protect\cite{zhu2015long}. \textbf{LSTM/BLSTM/Tree-LSTM}: Improved semantic representations from tree-structured long short-term memory networks \protect\cite{tai2015improved}. \textbf{LSTMN}: Long short-term memory-networks for machine reading \protect\cite{cheng2016long}. \textbf{Multi-Task}: Recurrent Neural Network for Text Classification with Multi-Task Learning \protect\cite{liu2016recurrent}. \textbf{PV}: Distributed representations of sentences and documents \protect\cite{le2014distributed}. \textbf{DAN}: Deep unordered composition rivals syntactic methods for text classification \protect\cite{iyyer2015deep}. \textbf{combine-skip}: skip-thought vectors \protect\cite{kiros2015skip}. \textbf{AdaSent}: Self-adaptive hierarchical sentence model \protect\cite{zhao2015self}. \textbf{LSTM-RNN}: Compositional distributional semantics with long short term memory \protect\cite{le2015compositional}. \textbf{C-LSTM}: A C-LSTM Neural Network for Text Classification \protect\cite{zhou2015c}. \textbf{DSCNN}: Dependency Sensitive Convolutional Neural Networks for Modeling Sentences and Documents \protect\cite{zhang2016dependency}.} \end{table*}


\section{Results}
\subsection{Overall Performance}
This work implements four models, BLSTM, BLSTM-Att, BLSTM-2DPooling, and BLSTM-2DCNN. Table 2 presents the performance of the four models and other state-of-the-art models on six classification tasks. The BLSTM-2DCNN model achieves excellent performance on 4 out of 6 tasks. Especially, it achieves $52.4\%$ and $89.5\%$ test accuracies on SST-1 and SST-2 respectively.


BLSTM-2DPooling performs worse than the state-of-the-art models. While we expect performance gains through the use of 2D convolution, we are surprised at the magnitude of the gains. BLSTM-CNN beats all baselines on SST-1, SST-2, and TREC datasets. As for Subj and MR datasets, BLSTM-2DCNN gets a second higher accuracies. 
Some of the previous techniques only work on sentences, but not paragraphs/documents with several sentences. Our question becomes whether it is possible to use our models for datasets that have a substantial number of words, such as 20Ng and where the content consists of many different topics. For that purpose, this paper tests the four models on document-level dataset 20Ng, by treating the document as a long sentence. Compared with RCNN \cite{lai2015recurrent}, BLSTM-2DCNN achieves a comparable result.

Besides, this paper also compares with ReNN, RNN, CNN and other neural networks:
\begin{itemize}
\item{Compared with ReNN, the proposed two models do not depend on  external language-specific features such as dependency parse trees.}

\item{CNN extracts features from word embeddings of the input text, while BLSTM-2DPooling and BLSTM-2DCNN captures features from the output of BLSTM layer, which has already extracted features from the original input text.}

\item{BLSTM-2DCNN is an extension of BLSTM-2DPooling, and the results show that BLSTM-2DCNN can capture more dependencies in text.}

\item{AdaSent utilizes a more complicated model to form a hierarchy of representations, and it outperforms BLSTM-2DCNN on Subj and MR datasets. Compared with DSCNN \cite{zhang2016dependency}, BLSTM-2DCNN outperforms it on five datasets.}
\end{itemize}

Compared with these results, 2D convolution and 2D max pooling operation are more effective for modeling sentence, even document. To better understand the effect of 2D operations, this work conducts a sensitivity analysis on SST-1 dataset.

\begin{figure}
\begin{tabular}{lr}
\begin{minipage}[t]{0.45\linewidth}
\centering
\includegraphics[scale=0.4]{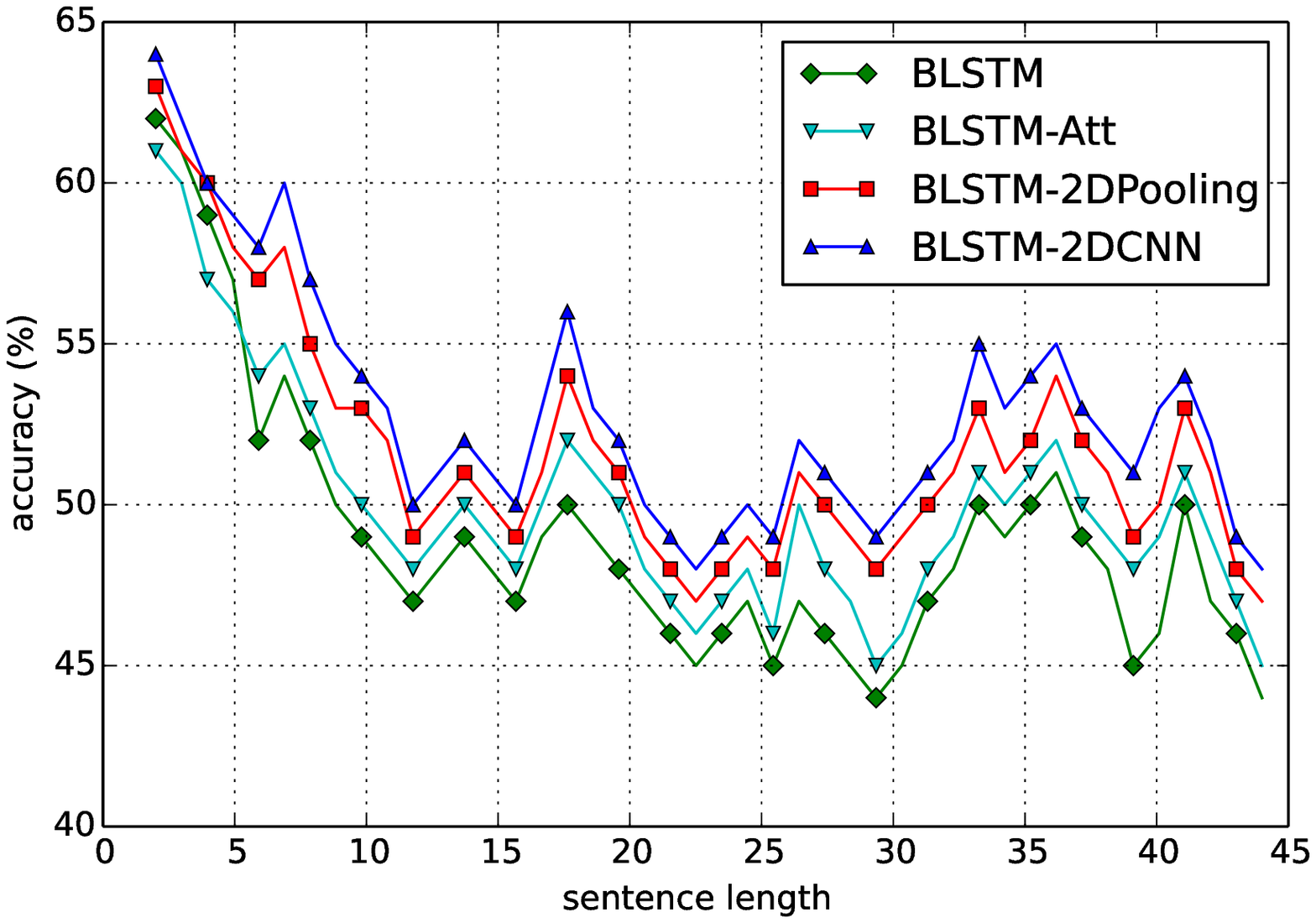}  
\caption{Fine-grained sentiment classification accuracy $vs.$ sentence length.}
\label{pool}
\end{minipage}
\hspace{0.2in}
\begin{minipage}[t]{0.45\linewidth}
\centering
\includegraphics[scale=0.4]{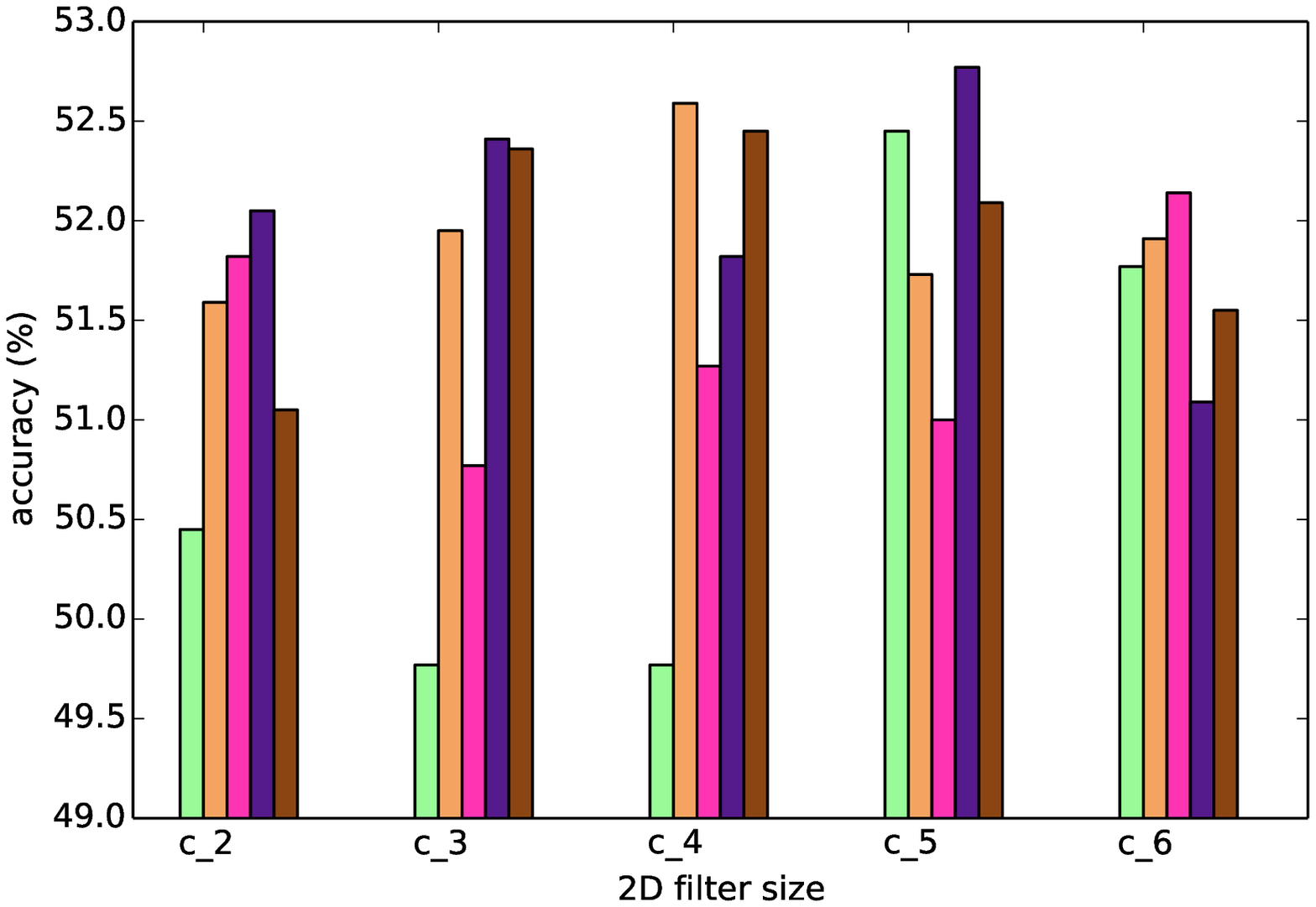}  
\caption{Prediction accuracy with different size of 2D filter and 2D max pooling.}
\label{pool}
\end{minipage}
\end{tabular}
\end{figure}


\subsection{Effect of Sentence Length}
Figure 2 depicts the performance of the four models on different length of sentences. In the figure, the x-axis represents sentence lengths and y-axis is accuracy. The sentences collected in test set are no longer than 45 words. The accuracy here is the average value of the sentences with length in the window $[l-2, l+2]$. Each data point is a mean score over 5 runs, and error bars have been omitted for clarity.

It is found that both BLSTM-2DPooling and BLSTM-2DCNN outperform the other two models. This suggests that both 2D convolution and 2D max pooling operation are able to encode semantically-useful structural information. At the same time, it shows that the accuracies decline with the length of sentences increasing. In future work, we would like to investigate neural mechanisms to preserve long-term dependencies of text.


\subsection{Effect of 2D Convolutional Filter and 2D Max Pooling Size}
We are interested in what is the best 2D filter and max pooling size to get better performance. We conduct experiments on SST-1 dataset with BLSTM-2DCNN and set the number of feature maps to 100.



To make it simple, we set these two dimensions to the same values, thus both the filter and the pooling are square matrices. For the horizontal axis, c means 2D convolutional filter size, and the five different color bar charts on each c represent different 2D max pooling size from 2 to 6. Figure 3 shows that different size of filter and pooling can get different accuracies. 
The best accuracy is 52.6 with 2D filter size (5,5) and 2D max pooling size (5,5), this shows that finer tuning can further improve the performance reported here. And if a larger filter is used, the convolution can detector more features, and the performance may be improved, too. However, the networks will take up more storage space, and consume more time. 

\section{Conclusion}
This paper introduces two combination models, one is BLSTM-2DPooling, the other is BLSTM-2DCNN, which can be seen as an extension of BLSTM-2DPooling. Both models can hold not only the time-step dimension but also the feature vector dimension information. The experiments are conducted on six text classificaion tasks. The experiments results demonstrate that BLSTM-2DCNN not only outperforms RecNN, RNN and CNN models, but also works better than the BLSTM-2DPooling and DSCNN \cite{zhang2016dependency}. Especially, BLSTM-2DCNN achieves highest accuracy on SST-1 and SST-2 datasets. To better understand the effective of the proposed two models, this work also conducts a sensitivity analysis on SST-1 dataset. It is found that large filter can detector more features, and this may lead to performance improvement. 


\section*{Acknowledgements}
We thank anonymous reviewers for their constructive comments. This research was funded by the National High Technology Research and Development Program of China (No.2015AA015402), and the National Natural Science Foundation of China (No. 61602479), and the Strategic Priority Research Program of the Chinese Academy of Sciences (Grant No. XDB02070005).

\bibliographystyle{acl}
\bibliography{coling2016}

\end{document}